\documentclass{article}

\usepackage[accepted]{icml2020_style/icml2020}

\usepackage[utf8]{inputenc} 
\usepackage[T1]{fontenc}    
\usepackage[pagebackref=true]{hyperref}       
\usepackage{url}            
\usepackage{booktabs}       
\usepackage{amsfonts}       
\usepackage{nicefrac}       
\usepackage{microtype}      
\usepackage{amsmath, amssymb}
\usepackage{graphicx}
\usepackage{caption}
\usepackage{subcaption}
\usepackage{xcolor}

\icmltitlerunning{Predicting Sim-to-Real Transfer with Probabilistic Dynamics Models}

\begin{document}

\twocolumn[
\icmltitle{Predicting Sim-to-Real Transfer of RL Policies \\ with Probabilistic Dynamics Models}



\icmlsetsymbol{equal}{*}

\begin{icmlauthorlist}
\icmlauthor{Lei M.~Zhang}{dm}
\icmlauthor{Matthias Plappert}{oai}
\icmlauthor{Wojciech Zaremba}{oai}
\end{icmlauthorlist}

\icmlaffiliation{dm}{DeepMind, Montr\'{e}al, QC, Canada. Work done while the author was a Fellow at OpenAI.}
\icmlaffiliation{oai}{OpenAI, San Francisco, CA, USA}

\icmlcorrespondingauthor{Lei M.~Zhang}{lmzhang@google.com}

\icmlkeywords{Machine Learning, ICML}

\vskip 0.3in
]



\printAffiliationsAndNotice{}

\begin{abstract}
We propose a method to predict the sim-to-real transfer performance of RL policies. Our transfer metric simplifies the selection of training setups (such as algorithm, hyperparameters, randomizations) and policies in simulation, without the need for extensive and time-consuming real-world rollouts. A probabilistic dynamics model is trained alongside the policy and evaluated on a fixed set of real-world trajectories to obtain the transfer metric. Experiments show that the transfer metric is highly correlated with policy performance in both simulated and real-world robotic environments for complex manipulation tasks. We further show that the transfer metric can predict the effect of training setups on policy transfer performance.
\end{abstract}

\section{Introduction}\label{sec:introduction}
Many recent advances in deep reinforcement learning (RL) for robotics have been achieved by training in simulation and zero (or few) shot transfer to a real environment \cite{Tan2018, Peng2018, Golemo2018, OpenAI2018, Hwangbo2019, OpenAI2019}. The key advantage of this approach is the availability of efficient and scalable environment interactions in simulation and is well-suited to powerful deep RL algorithms \cite{Schulman2017, Haarnoja2018}. However, policies trained in simulation often do not achieve the same level of performance in the real environment. This is known as the sim-to-real transfer problem.

Domain randomization (DR) \cite{Tobin2017, Sadeghi2017, Peng2018} is a promising technique which helps to resolve the sim-to-real transfer problem. In DR, a policy is trained using data generated from a diverse set of environments. The environments are sampled from a distribution defined by a set of manually tuned parameters \cite{Tobin2017, Sadeghi2017, Peng2018, OpenAI2018}. Success of DR depends on the parameter tuning, which involves iterating between policy training and real-world deployment. This iterative process is slow, prone to failures, and human dependent.

Automatic domain randomization (ADR) \cite{OpenAI2019} improves on DR by automatically adjusting the environment distribution based on policy performance. However, ADR suffers from a similar iterative process as DR, since it is difficult to know when the environment distribution is sufficient to produce good transfer.

In this work, we introduce a method to predict the transfer performance of policies trained in simulation, using only a fixed set of real-world trajectories. We show that the proposed \emph{transfer metric} strongly correlates with policy transfer performance. For DR training, we show that the transfer metric can predict the transfer performance of policies trained using different environment distribution parameters. For ADR training, we show that the transfer metric can predict relative changes in transfer performance as training progresses and can be used as an effective indicator for stopping ADR.

\section{Related Works}\label{sec:related}
A number of methods that use real-world data to improve sim-to-real performance have been proposed. Neural networks were trained to predict and compensate for discrepancies between simulated and real-world trajectories \cite{Golemo2018, Hwangbo2019}. An inverse dynamics model was trained using real-world rollouts and used to adjust policy outputs during deployment \cite{Christiano2016}. A policy network was trained in simulation then progressively modified in the real-world environment to improve performance \cite{Rusu2016}. These methods rely on a large amount of real-world trajectories to train neural networks, limiting their scope of application.

Works on predicting transfer performance have been suggested for evolution strategies \cite{Koos2010} with evaluation on a fixed set of real-world trajectories \cite{Farchy2013, Hanna2017}. The grounded simulation learning (GSL) technique proposed in \cite{Hanna2017} optimizes the simulator based on its predictions on real-world trajectories. Our work implements a similar idea by using a more flexible (and differentiable) dynamics model.

For deep RL policies trained under DR, \cite{Muratore2018} proposed a method of assessing transfer performance based on upper-confidence bounds with significant computational requirements. \cite{Chebotar2018} defined a discrepancy between simulation and real-world trajectories then tuned DR probability distribution based on the discrepancy. In \cite{Mehta2019}, a discriminator was trained to distinguish between simulated and real trajectories and used to tune the DR probability distribution. Similar methods have also been used to improve transfer performance of vision models \cite{Prakash2018, Pashevich2019}. 

More recently, \cite{Irpan2019} used a learned value function and real-world trajectories to predict transfer performance of deterministic policies in binary reward, deterministic environments. 

\section{Method}
\subsection{Definitions}
We define a environment to be a Markov decision processes (MDP) specified by a tuple $(S, A, R, p, p_0, \gamma)$, where $S$ is the state space, $A$ is the action space, $R: S \times A \mapsto \mathbf{R}$ is the reward function, $p(s_{t+1} | s_t, a_t): S \times S \times A \mapsto \mathbf{R}^{+}$ are the state transition probabilities with initial state distribution $p_0: S \mapsto \mathbf{R}^{+}$, and $\gamma \in [0, 1]$ is the discount factor.

If an MDP is partially-observable (POMDP), the above definitions are modified by an additional observation function $o: S \mapsto O$ used to map states to observations. Our method extends naturally to an POMDP and we refer to observations $o_t$ in a POMDP as states denoted by $s_t$ to avoid unnecessary notation.

A \emph{source environment} is an environment that is used to train a policy. Under DR and ADR, a distribution of source environments is used. All source environments in this work are simulations based on the MuJoCo simulator \cite{Todorov2012}.

A \emph{target environment} is an environment in which a policy is deployed after training. When a target environment is simulated and different from source environments, it is referred to as a \emph{held-out environment}. We also consider real-world target environments, such as the Shadow robotic hand \cite{OpenAI2018} in this work.

A \emph{trajectory} is a discrete-time sequence of states, actions, and rewards
\[
\tau \triangleq (s_0, a_0, r_0, \dots, s_{T-1}, a_{T-1}, r_{T-1}, s_T),
\]
where $T$ is the terminal time step.

\subsection{Transfer Metric Definition}
We first train a prediction model in the same source environment (or environments) used to train our policy. We then evaluate its prediction accuracy on a fixed set of trajectories from the target environment. The transfer metric is defined as the resulting prediction accuracy.

A key assumption is that the model prediction accuracy in the target environment correlates strongly with the performance of our policy in the same environment. To make this more likely, we enforced the same architecture, hyperparameters, and training data when training both networks. Experiments in Sec.~\ref{sec:experiments} show that such strong correlations indeed exist.

A key benefit of the transfer metric is that it does not require expensive rollouts of the trained policy in the target environment. It does require a set of pre-recorded trajectories from the target environment, which can be recorded using a sub-optimal behavioral policy.
\begin{figure}[t]
\centering
\includegraphics[width=0.9\linewidth]{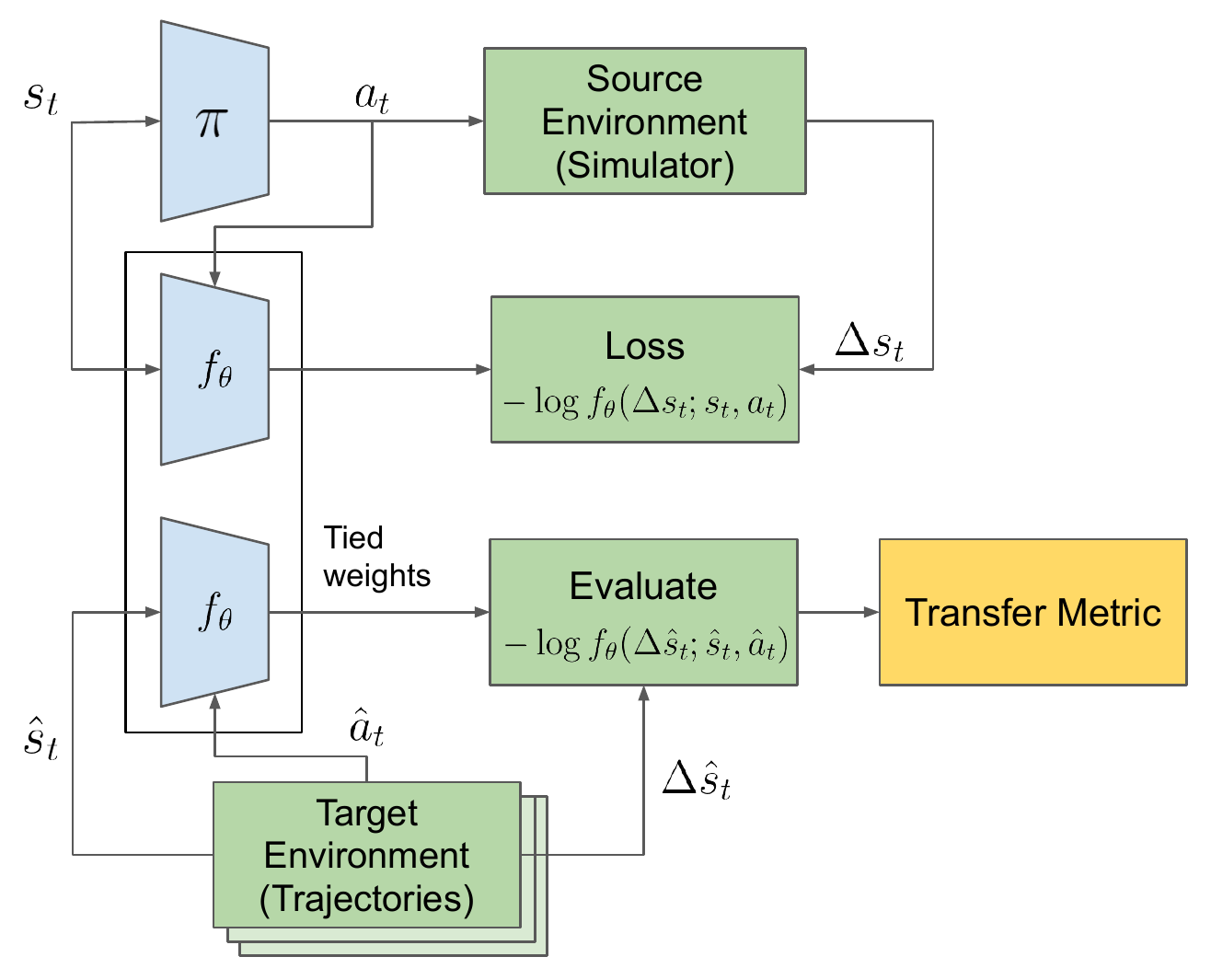}
\caption{Transfer metric architecture. A probabilistic dynamics model is trained in the same source environment as the policy. The transfer metric is the average negative log-likelihood of the relative next state given by the model for trajectories from the target environment.}
\label{fig:transfer_metric_top}
\end{figure} 

Several aspects of an environment may be predicted by the model: state, action\footnote{We also tried action prediction by using inverse dynamics models. We found that it had slightly worse correlation with transfer performance than state prediction.}, reward, and return \cite{Irpan2019}. We chose to predict the relative next state $\Delta s_t \triangleq s_{t+1} - s_t \in S'$ from current state and action. We found relative state prediction to work better than absolute state prediction, which is consistent with prior work on training forward dynamics models \cite{Deisenroth2015, Chua2018}. A top-level diagram of the transfer metric is shown in Fig.~\ref{fig:transfer_metric_top}.

We used a probabilistic forward dynamics model $f_{\theta}(\Delta s_t; s_t, a_t) = P(\Delta s_t | s_t, a_t)$, where $\theta$ were the model parameters. For POMDPs, the inputs to the model were sequences of states and actions.

The prediction space $S'$ was quantized to a set of discrete values. We found this helped model training significantly. The forward dynamics model was trained in the source environment using a negative log-likelihood (NLL) loss.

The transfer metric is the average NLL of the forward dynamics model evaluated on a set of pre-recorded trajectories from the target environment.

\section{Experiments}\label{sec:experiments}
\subsection{Training Setup and Benchmark Tasks}\label{sub:setup}
Our experiments used the same policy and value function architectures as in \cite{OpenAI2019}. Each consisted of input embeddings, feed-forward layers, an LSTM core, and an output layer. PPO \cite{Schulman2017} or SAC \cite{Haarnoja2018a} with hindsight experience replay (HER) \cite{Andrychowicz2017} was used for policy training.

We used the following benchmark dexterous in-hand manipulation tasks:
\paragraph{Block reorientation task} \cite{OpenAI2018, OpenAI2019} requires the robot hand to orient a solid cube to match a goal orientation. Once the orientation matches, a successful rotation is declared and a new goal orientation is obtained. An episode continues until the robot hand either drops the cube or achieves 50 successful rotations.

\paragraph{Rubik's cube task} \cite{OpenAI2019} requires the robot to either rotate the top face of a Rubik's cube by 90 degrees to match a goal face orientation or orient the entire Rubik's cube to match a goal cube orientation. Here, we do not distinguish between face and cube rotations and declare a successful rotation when either occurs. The same episode termination conditions apply.

\subsection{Simulator Randomizations}
Domain randomization was used for policy training in simulation. The randomizations fall into two categories:

We apply \emph{parameter randomizations} to environment parameters in the simulator. A calibration value is specified for a parameter, along with a fixed (in DR) or adjusted (in ADR) range centered at the calibration value. At the start of an episode, every randomized parameter is sampled uniformly from its range.

We additionally apply \emph{custom physics randomizations} to incorporate real-world effects (such as motor backlash or action latency) into the simulation environment. We focus on five custom physics randomizations in particular: freezing cube, backlash, wind, action latency, and action noise. 

Freezing cube randomly prevents the cube from being moved and is intended to model the dynamics of a cube stuck in some parts of the hand. Backlash models the motor backlash effect, as described in detail in \cite{OpenAI2018}. Wind models air flow around the cube. Action latency and noise are modifies how an action command from a policy is executed in simulation.

Each customs physics randomization involves some custom implementation of the physical dynamics with a set of parameters. A subset of these parameters are then randomized in a similar fashion as parameter randomization. For more details, see \cite{OpenAI2018, OpenAI2019}.

\subsection{Sim-to-sim Transfer}\label{sub:sim2sim}
In these experiments:
\vspace{-1em}
\begin{itemize}
\setlength\itemsep{0em}
    \item Source Env: randomized simulation environments
    \item Target Env: held-out simulation environments
\end{itemize}
\vspace{-0.5em}

We used three parameter randomization methods in the held-out simulation environments:
\vspace{-1em}
\begin{itemize}
\setlength\itemsep{0em}
\item \texttt{cal}: always use the calibration value
\item \texttt{ext}: uniformly sample the upper or lower limit of the parameter range
\item \texttt{rand}: uniformly sample within the parameter range
\end{itemize}
\vspace{-0.5em}

We additionally used three different sets of custom physics randomizations in the held-out simulation environments:
\vspace{-1em}
\begin{itemize}
\setlength\itemsep{0em}
\item \texttt{none}: no custom physics
\item \texttt{partial}: action latency, action noise
\item \texttt{all}: freezing cube, backlash, wind, action latency, action noise
\end{itemize}
\vspace{-0.5em}

\begin{figure}
\centering
\includegraphics[width=\linewidth]{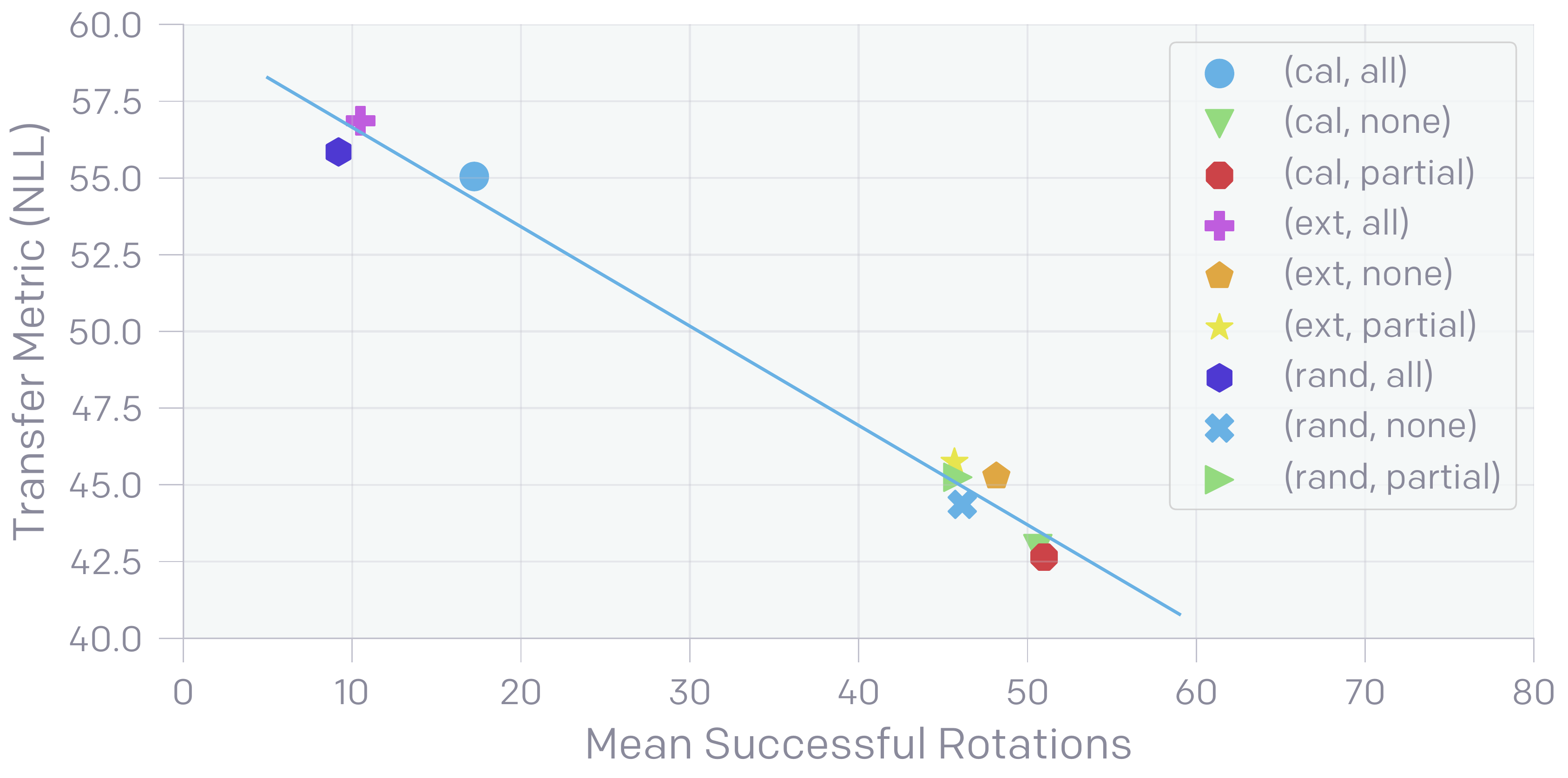}
\caption{Transfer metric (NLL, lower is better) vs. performance in held-out environments. Training was performed in environments with (\texttt{rand}, \texttt{partial}) randomizations. The legend refers to the parameter and custom physics randomizations used in the held-out environment. The linear best-fit line has coefficient of determination of $R^2=-0.99$. Transfer metric strongly correlates with policy performance in held-out environments.}
\label{fig:sim2sim_correlation}
\end{figure}

The set of all nine possible held-out environments given by is $\{\texttt{cal}, \texttt{ext}, \texttt{rand}\} \times \{\texttt{none}, \texttt{partial}, \texttt{all}\}$. We focused on the block reorientation task in these experiments. The source environments used for policy training is (\texttt{rand}, \texttt{partial}). Policy optimization was based on SAC with HER. We evaluated the prediction model on held-out environments directly to obtained the transfer metric. Fig.~\ref{fig:sim2sim_correlation} shows a plot of the transfer metric and held-out performance.

The results show that the transfer metric predicts policy performance in a diverse set of held-out environments with strong linear correlation. It was also able to predict the effect of progressively adding more custom physics randomizations, for example the transfer metric increases quickly between (\texttt{ext}, \texttt{none}), (\texttt{ext}, \texttt{partial}), and (\texttt{ext}, \texttt{all}).

\subsection{Sim-to-Real Transfer}\label{sub:sim2real}
In these experiments:
\vspace{-1em}
\begin{itemize}
\setlength\itemsep{0em}
    \item Source Env: randomized simulation environments
    \item Target Env: physical Shadow robotic hand
\end{itemize}
\vspace{-0.5em}
We performed real-world deployments of policies for both the block reorientation task and the Rubik's cube task. For block reorientation, we evaluated transfer metrics on a fixed set of trajectories recorded from Shadow robotic hands. They contains approximately 500 trajectories ranging from 20 to 3000 time steps. See Sec.~\ref{sub:rubiks_cube} for details of the recorded trajectories used to evaluate the transfer metrics for the Rubik's cube task.

\subsubsection{Block Reorientation Policies}
\begin{figure}[h]
    \centering
    \begin{subfigure}[b]{\linewidth}
	\includegraphics[width=\linewidth]{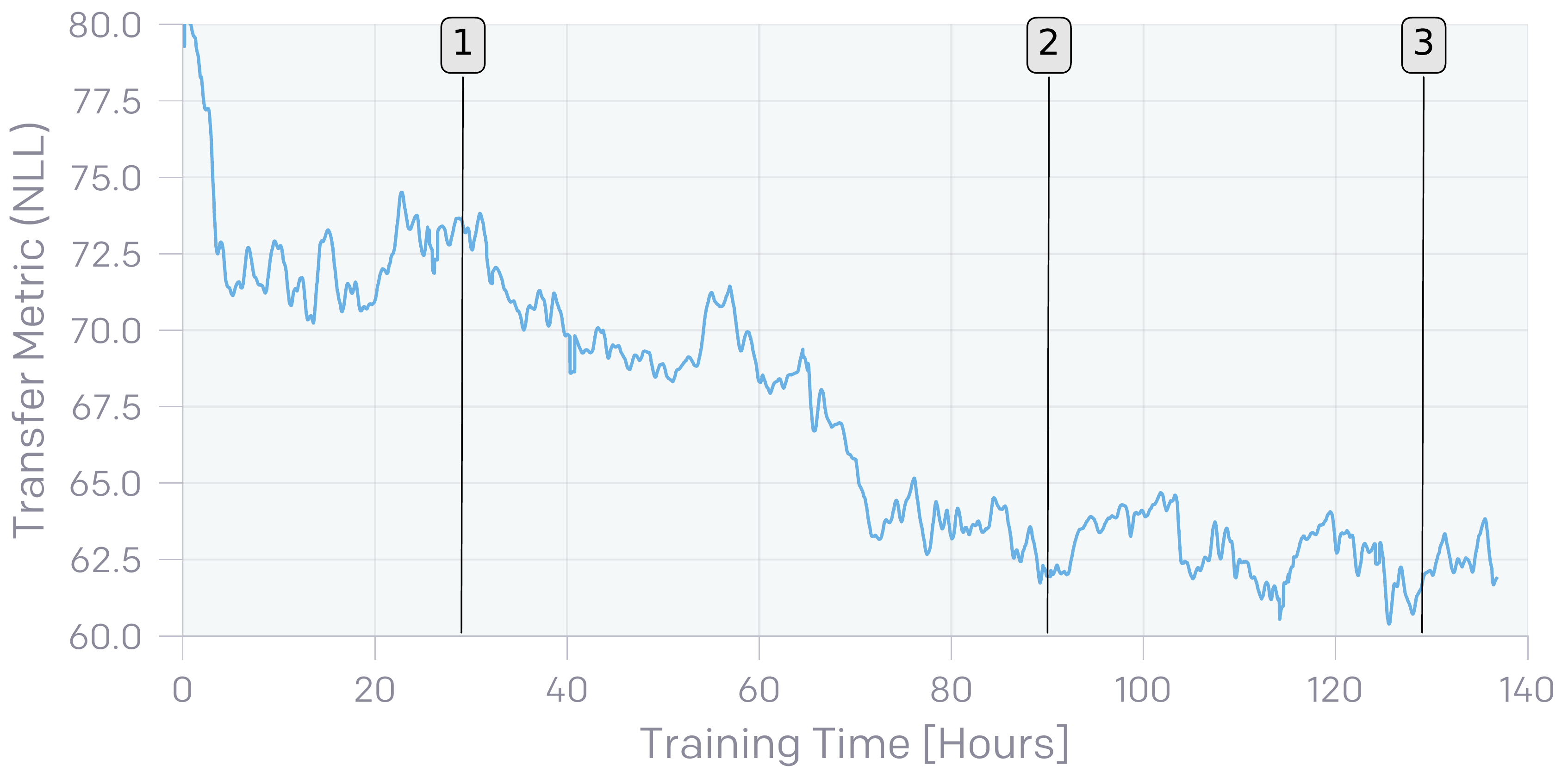}
        \caption{Transfer Metric}
    \end{subfigure}
    \\
    \begin{subfigure}[b]{\linewidth}
	\includegraphics[width=\linewidth]{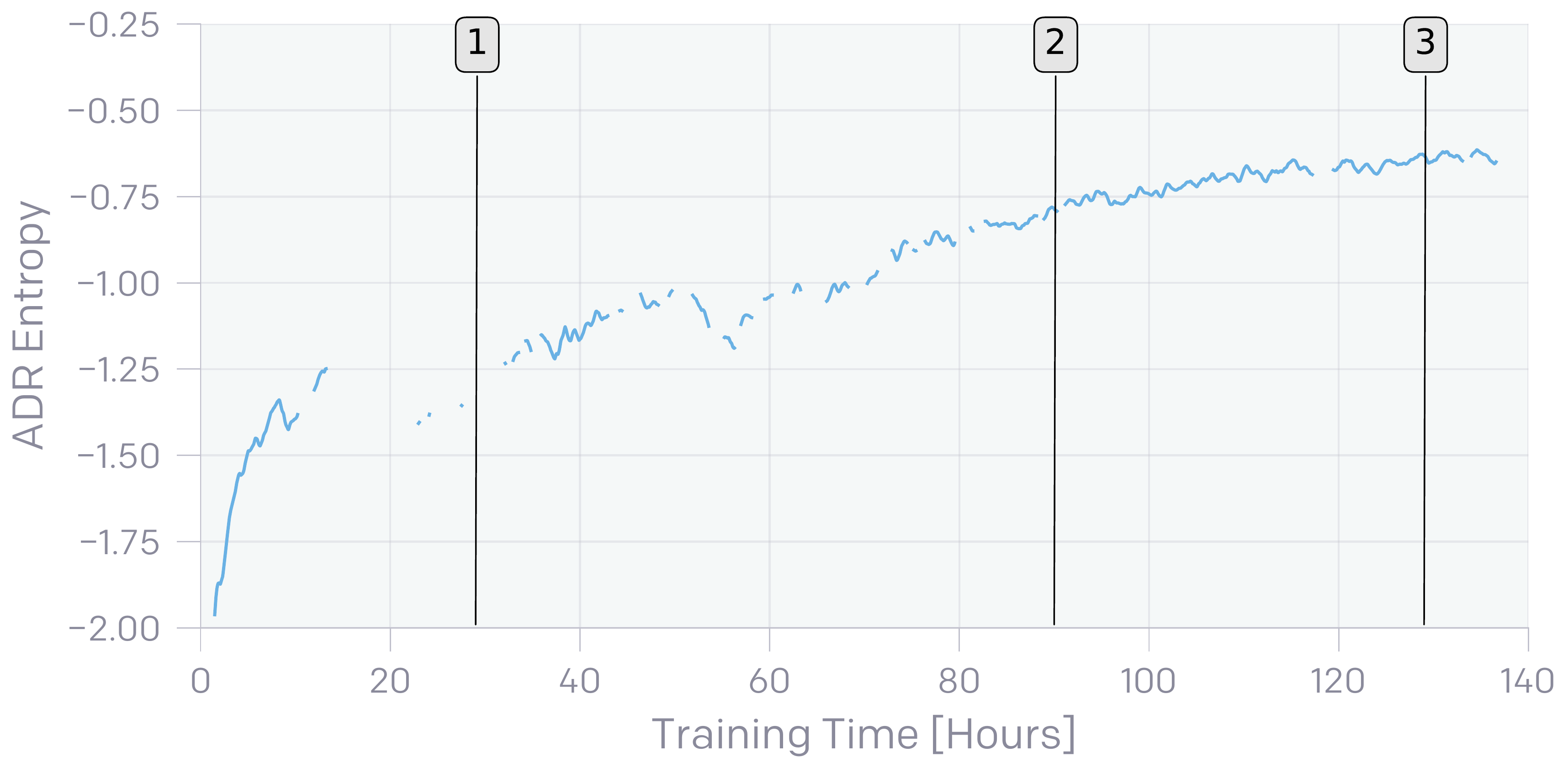}
        \caption{ADR Entropy}
    \end{subfigure}
    \caption{Transfer metric and ADR entropy of ADR Run 1 for block reorientation. Policies were taken at the marked training times and deployed on a real robot. Missing segments in ADR entropy are due to an entropy value of -$\infty$. Note the decrease and flattening out of the transfer metric across the three policies in contrast to the constantly increasing ADR entropy.}
    \label{fig:metric_entropy_compare}
\end{figure}

We picked three policies at different checkpoints of an ADR training run (`Run 1') for the block reorientation task. Fig.~\ref{fig:metric_entropy_compare} shows the transfer metric and ADR entropy throughout the ADR training run. ADR entropy is the differential entropy of the environment distribution normalized by the number of environment parameters. It has been shown to correlate with real-world performance, where a higher ADR entropy is associated with better performance \cite{OpenAI2019}. As we show below, this is not true in general.

We picked the policies so that there is a large change in both transfer metric and ADR entropy between policies 1 and 2 but a nearly constant transfer metric between policies 2 and 3 while ADR entropy continues to increase.

We deployed the policies on a robotic hand by interleaving 5 episodes of each policy until 20 episodes have been collected for every policy. This procedure ensures that the robot condition is consistent across all policies. The results are shown in Table~\ref{tbl:transfer_1}. For each ADR run (first column), the p-value is calculated using the two-sided, two-sample unequal variance t-test between the means of consecutive policies (with wrap-around), with the null hypothesis being equal means.

As expected, the results show a significant improvement in performance between policies 1 and 2. This is not surprising since both the transfer metric and ADR entropy changed significantly between these policies (not to mention the longer training time). 

More interesting result is the small change in performance between policies 2 and 3. This is surprising when judged by ADR entropy or training time, which both increased between policies 2 and 3 and at the same rate as between policies 1 and 2. The transfer metric correctly predicted that performance will not continue to improve.

\begin{table*}[t]
\caption{Performance of block reorientation policies on robotic hand. Policies were trained under ADR with SAC. An ADR entropy of $-\infty$ corresponds to one or more components of the environment distribution being set to zero, see \cite{OpenAI2019}. ADR entropy is in units of nats-per-dimension (npd). For run 1, policy checkpoints refer to the markers in Fig.~\ref{fig:metric_entropy_compare}}
\label{tbl:transfer_1}
\centering
\begin{tabular}{ccccrcc}
\toprule
&&&\multicolumn{4}{c}{Number of successful rotations} \\
\cmidrule(r){4-7} 
Run & Policy Checkpoint & Time (days) & Transfer metric & ADR entropy & Mean & p-value \\
\midrule
1 & 1 & 0.74 & 80.7 & $-\infty$ npd & 1.55 & 0.022 \\
 & 2 & 3.8 & 61.6 & $-0.78$ npd & 3.35 & 0.54 \\
 & 3 & 5.4 & 59.4 & $-0.65$ npd & 3.95 & 0.006 \\
\midrule
2 & 1 & 4.3 & 58.6 & $-\infty$ npd & 3.75 & 0.96 \\
 & 2 & 5.5 & 58.8 & $-0.51$ npd & 3.70 & 0.96 \\
\midrule
3 & 1 & 2.5 & 73.7 & $-\infty$ npd & 1.90 & 0.094 \\
 & 2 & 5.2 & 59.0 & $-0.81$ npd & 3.15 & 0.43 \\
 & 3 & 5.8 & 63.0 & $-0.73$ npd & 2.55 & 0.26 \\
\bottomrule
\end{tabular}
\end{table*}

As shown in Table~\ref{tbl:transfer_1}, we repeated the above experiment for two additional ADR training runs: `Run 2' and `Run 3'. The policies for `Run 2' were picked to have essentially the same transfer metric ($58.6$ and $58.8$) but different ADR entropy. They are also separated in training time by more than a day. Performance of these policies on the robot were virtually identical.

The policies for `Run 3' selected with a similar intention as `Run 1'. The difference here is that the transfer metric between policies 2 and 3 predicts a performance \emph{decrease} while ADR entropy continues to increase. The results show that indeed, performance decreased from $3.15$ to $2.55$ mean successful rotations.

In summary, we showed in these experiments that the transfer metric is predictive of the block reorientation policy performance on a real-world robot. We further showed that the predictions cannot be explained by longer training times and are more accurate than ADR entropy.

\subsubsection{Rubik's Cube Policies}\label{sub:rubiks_cube}
In these experiments, we study the sensitivity of the transfer metric to changes on the real-world robot. We performed 8 deployments of the \emph{same} policy for the Rubik's cube task. The policy was trained under ADR with PPO.

Deployments were performed on three different robotic hands, in two different state sensing setups (`cages') \cite{OpenAI2019} and with varying amounts of mechanical and calibration degradation. The mean number of successful rotations of these deployments ranged from $4.3$ to $20.3$. Each deployment was recorded to be used for the transfer metric.

\begin{figure}
\centering
\includegraphics[width=\linewidth]{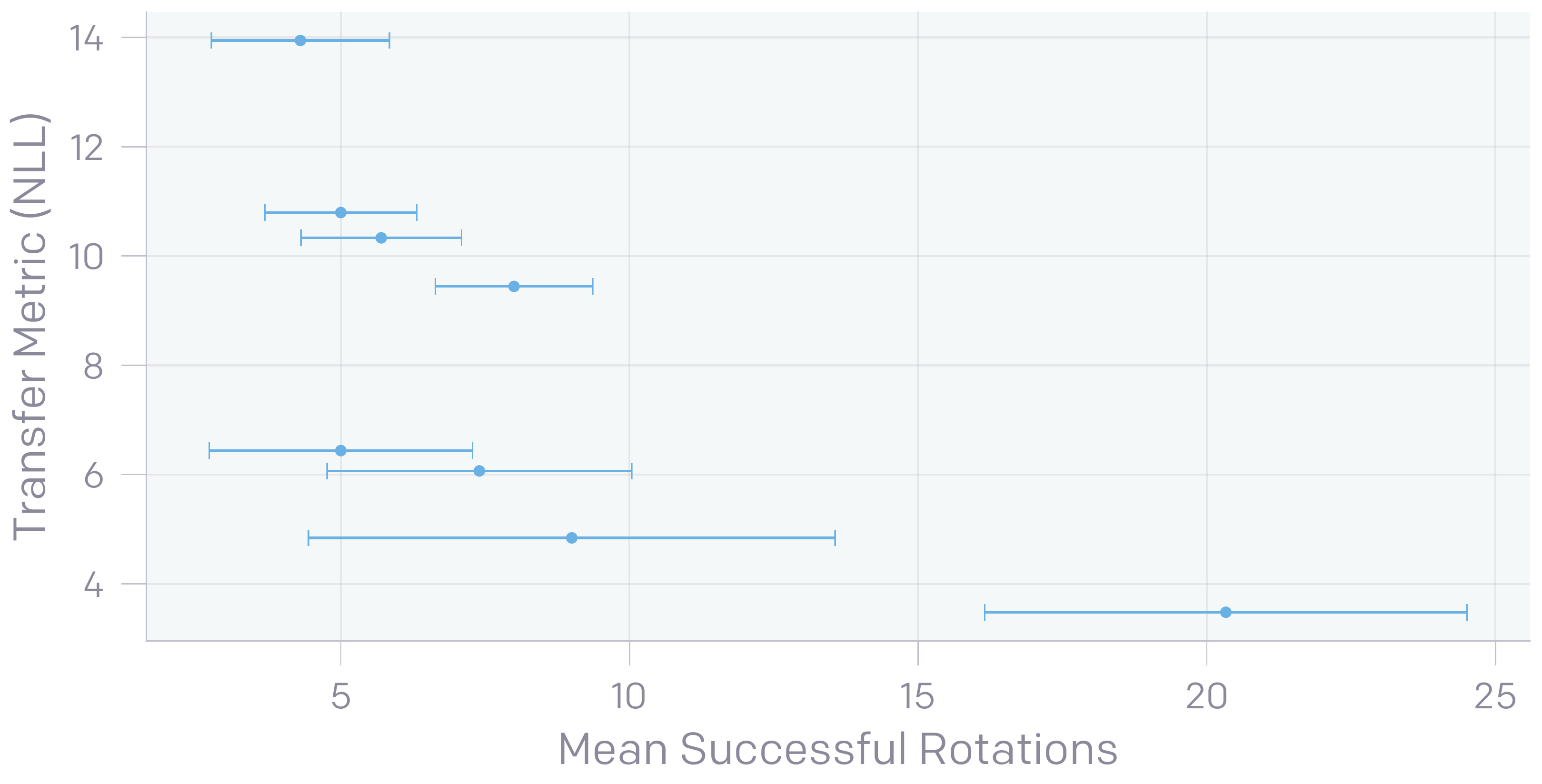}
\caption{Performance and transfer metric of a single Rubik's cube policy on different robotic hand setups (with different state-sensing setups and varying amounts of mechanical and calibration degradation). Error bars mark the standard error of the mean.}
\label{fig:rubiks_cube_plot}
\end{figure}

We evaluated a transfer metric for each deployment using its recorded trajectory. The results are shown in Fig.~\ref{fig:rubiks_cube_plot}. We see that the transfer metric is able to predict the ordering of the performance on the robot. 

In summary, we showed in these experiments that the transfer metric is sensitive to changes in the robotic hand. We have shown this for the Rubik's cube task, a much harder manipulation task than block reorientation. 

\subsubsection{Transfer Metrics of Existing Policies}
In these experiments, we evaluated the transfer metric for policies that were deployed on the robotic hand and have available performance data. For each existing policy, we added a prediction model and performed rollouts in simulation. We optimized only the prediction model parameters using the rollout data. If ADR was used to train the policy, we also froze the ADR updates.

Table~\ref{tbl:block_reorientation_old} shows the transfer metrics for existing block reorientation policies. The policies were originally trained under ADR with PPO or SAC. The transfer metrics were evaluated on the same block reorientation recordings as in the previous sections.

\begin{table}
\caption{Transfer metrics for existing block reorientation policies. The data contains a mix of policies trained with PPO or SAC under ADR. Different prediction architectures were used when training using PPO or SAC.}
\label{tbl:block_reorientation_old}
\centering
\begin{tabular}{cc p{1.5cm} p{2.0cm}}
\toprule
Policy & Algorithm & Transfer metric  & Mean successful rotations \\
\midrule
1 & SAC & 59.5 & 16.5 \\
2 & SAC & 68.8 & 4.4 \\
3 & SAC & 68.5 & 2.6 \\
4 & SAC & 55.4 & 17.3 \\
5 & SAC & 59.4 & 30.7 \\
\midrule
1 & PPO & 46.3 & 17.2 \\
2 & PPO & 43.6 & 15.7 \\
3 & PPO & 42.3 & 20.4 \\
\bottomrule
\end{tabular}
\end{table}

Good correlation is seen between the transfer metric and performance of existing block reorientation policies for the same algorithm.

Table~\ref{tbl:rubiks_cube_old} shows the transfer metrics for existing Rubik's cube policies \cite{OpenAI2019}. These policies were trained using PPO under ADR. The transfer metrics were evaluated on the same Rubik's cube recordings as in the previous sections.

\begin{table}
\caption{Transfer metrics for existing Rubik's cube policies. ADR entropy is in units of nats-per-dimension (npd).}
\label{tbl:rubiks_cube_old}
\centering
\begin{tabular}{c p{1.7cm} p{1.5cm} p{2.0cm}}
\toprule
Policy & ADR entropy & Transfer metric  & Mean successful rotations \\
\midrule
1 & $-0.054$ npd & 73.6 & 3.8 \\
2 & $0.467$ npd & 68.0 & 17.8 \\
3 & $0.479$ npd & 67.2 & 26.8 \\
\bottomrule
\end{tabular}
\end{table}

Again, good correlation is seen between the transfer metric and performance of existing Rubik's cube policies.

In summary, we showed in these experiments that it is not necessary to train the prediction model alongside the policy and that the transfer metric can be evaluated for existing policies. We also showed good correlation between the evaluate transfer metrics for existing both block reorientation and Rubik's cube policies.

\subsection{Transfer Metric and Domain Randomization}\label{sub:det_dr_adr}
In these experiments:
\vspace{-1em}
\begin{itemize}
\setlength\itemsep{0em}
    \item Source Env: randomized simulation environments
    \item Target Env: physical Shadow robotic hand\footnote{No rollouts, only evaluation on recorded trajectories.}
\end{itemize}
\vspace{-0.5em}
Here, we study the ability of the transfer metric to predict the effect of different levels of (fixed or automatic) domain randomization on policy transfer performance.

We trained block reorientation policies in \texttt{Cal}, \texttt{DR}, and \texttt{ADR} source environments. In the notation introduced in Sec.~\ref{sub:sim2sim}, the \texttt{Cal} source environment is (\texttt{cal}, \texttt{full}), where each parameter is fixed to a calibration value and all custom physics randomization are applied. The \texttt{DR} source environment is (\texttt{rand}, \texttt{full}), where each parameter is sampled uniformly from a fixed distribution and all custom physics randomizations are applied. The \texttt{ADR} source environment is the same as introduced in \cite{OpenAI2019}.

We evaluated the transfer metric on the set of trajectories given in Sec.~\ref{sub:sim2real}. In these rollouts, the average performance of policies trained under \texttt{Cal} source environments is 2.5 mean successful rotations. Average performance for policies trained under \texttt{DR} source environments is 5.6 and under \texttt{ADR} source environments is 25.0. Therefore, the rollout data indicate that training under \texttt{ADR} is much better in terms of transfer performance than \texttt{DR}, which is still better than \texttt{Cal}.

\begin{figure}
\centering
\includegraphics[width=\linewidth]{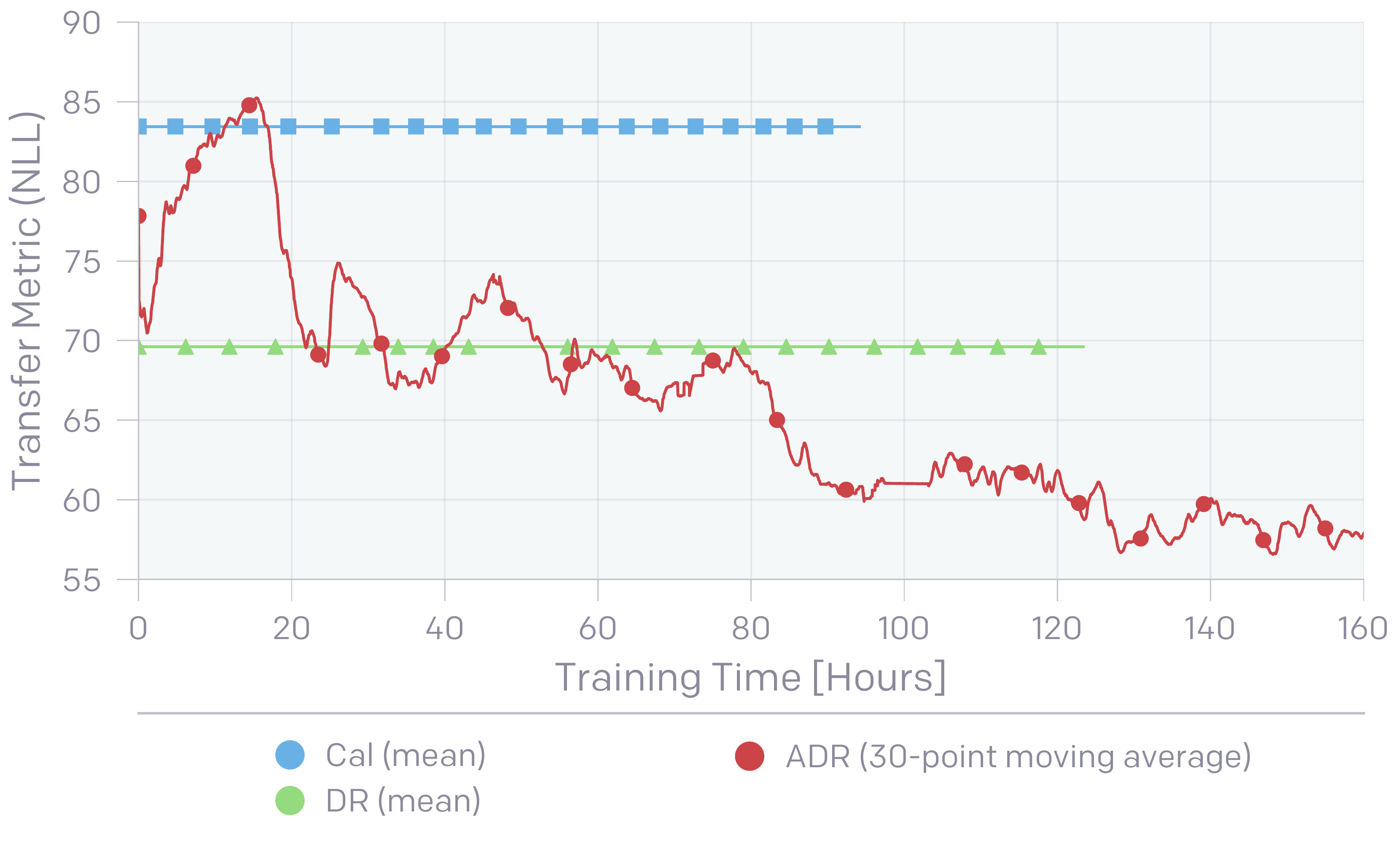}
\caption{Transfer metric (NLL, lower is better) of different source environments. Transfer metrics for \texttt{Cal} and \texttt{DR} are averaged over the entire training run. Transfer metrics for \texttt{ADR} are smoothed by 30-point moving average. The transfer metric predicts the correct ordering of transfer performance: $\texttt{ADR} > \texttt{DR} > \texttt{Cal}$.}
\label{fig:det_dr_adr}
\end{figure}

Fig.~\ref{fig:det_dr_adr} shows the transfer metric values for the three different sets of source environments. The transfer metric accurately predicts the relative ordering of average transfer performance of the resulting policies: $\texttt{ADR} > \texttt{DR} > \texttt{Cal}$. 

We note that the initial dip in the \texttt{ADR} transfer metric is also present in the other two cases (only their time-averages are shown in Fig.~\ref{fig:det_dr_adr}). This is likely due to the initial randomly initialized policy does not change the observation state too much, hence the next state is trivial to predict.

We also find that the \texttt{ADR} transfer metric evolves to first match the mean \texttt{Cal} transfer metric (after 18 hours of training), followed by the mean \texttt{DR} transfer metric (after 40 hours of training), then improves beyond both with additional training time. An explanation of this behaviour is that the environment distribution for ADR is initialized at the calibration environment, then automatically expands to approximate the manual DR distribution and beyond.

\section{Conclusion}
In this work, we proposed a method of predicting the real-world performance of policies trained in simulation, using only a fixed set of real-world trajectories. Our experiments showed that our transfer metric can successfully predict transfer performance in: 
\vspace{-1em}
\begin{itemize}
\setlength\itemsep{0em}
    \item Held-out simulation environments
    \item Real-world block reorientation with robotic hand
    \item Real-world Rubik's cube with robotic hand
    \item Existing policies for block reorientation and Rubik's cube
\end{itemize}
\vspace{-0.5cm}
We also showed that our transfer metric can be used to select levels of domain randomization for best transfer performance and is an effective criterion for stopping ADR training.

\bibliographystyle{icml2020_style/icml2020}
\bibliography{./bib/fellow}

\end{document}